\documentclass[11pt, twocolumn]{article}

\usepackage[authoryear, sort&compress, round]{natbib}
\usepackage{graphicx}
\usepackage[small]{caption}

\usepackage[hyphens]{url}
\usepackage[hidelinks]{hyperref}
\usepackage[hyphenbreaks]{breakurl}

\usepackage{geometry}

\usepackage{titlesec}
\usepackage{authblk}

\titleformat{\section}{\large\bfseries}{\thesection}{1em}{}

\title{Simulacra as Conscious Exotica}

\author[1,2,3]{Murray Shanahan \thanks{m.shanahan@imperial.ac.uk}}
\affil[1]{Google DeepMind}
\affil[2]{Imperial College London}
\affil[3]{Institute of Philosophy, School of Advanced Study, University of London}

\date{
February 2024\\
Updated July 2024
\vspace{\baselineskip}
\begin{flushleft}
\small{``[O]nly of a living human being and what resembles (behaves like) a living human being can one say: it has sensations ... is conscious or unconscious.'' Ludwig Wittgenstein, {\it Philosophical Investigations}}
\end{flushleft}
}

\begin{document}

\maketitle

\begin{abstract}
The advent of conversational agents with increasingly human-like behaviour throws old philosophical questions into new light. Does it, or could it, ever make sense to speak of AI agents built out of generative language models in terms of consciousness, given that they are ``mere'' simulacra of human behaviour, and that what they do can be seen as ``merely'' role play? Drawing on the later writings of Wittgenstein, this paper attempts to tackle this question while avoiding the pitfalls of dualistic thinking.
\end{abstract}

\section{Introduction}

As the behaviour of generative AI systems, especially conversational agents based on {\em large language models} (LLMs), becomes more compellingly human-like, so the temptation to ascribe human qualities to them will become increasingly hard to resist. The temptation will be especially great if those agents are {\em virtually embodied}, if they interact with users in immersive virtual worlds through avatars with a human- or animal-like form.\footnote{The term ``agent'' is used liberally throughout this paper. Although it has anthropomorphic overtones, it is used as a technical term in AI \citep[Chapter 2]{russell2010artificial} that doesn't imply agency in its fullest sense \citep{schlosser2019agenscy}.} Consciousness is one such quality, one that is especially morally valenced. To see a fellow creature as conscious (or sentient) often goes hand-in-hand with the sense that we should behave decently towards it \citep{singer1975animal, nussbaum2023justice}. To see an AI system as conscious would be to admit it into this fellowship of conscious beings, and potentially to give it ``moral standing'', which would be a serious matter \citep{metzinger2021artificial, ladak2023moralstanding}.

Unfortunately, the ascription of consciousness, except in the ordinary case of human beings, is a tricky business, and the less human-like the candidate, the more problematic the ascription \citep{shevlin2021nonhuman, mudrik2023theories, bayne2024tests}.\footnote{Even in humans, there are difficult clinical cases \citep{monti2010willful}.} Yet philosophical intuition is inclined to grant the possibility that, within the {\em space of possible minds} -- a space that encompasses not only humans and other animals, but also putative extraterrestrial life-forms as well as sufficiently advanced kinds of artificial intelligence \citep{sloman1984space} -- there exist {\em conscious exotica}, entities that are extremely different from anything found in (terrestrial) biology but that belong to the fellowship of conscious beings \citep{nagel1974what, shanahan2016conscious}.\footnote{The philosophical intuition in question manifests most prominently in the tropes of science fiction, but this is no reason to sideline it \citep{rennick2021trope}.}

LLM-based conversational agents are certainly exotic, in the relevant sense, notwithstanding their human-like behaviour. Their constitution is fundamentally different from a human's, or from that of any animal. Humans learn language through embodied interaction with other language users in a shared world, whereas a large language model is a disembodied computational entity that, at a fundamental level, predicts the next word (technically the next token) in a sequence of words (tokens), having been trained on a very large corpus of textual data \citep{bender2020climbing, shanahan2024talking}. As such, LLM-based conversational agents can be considered as {\em simulacra} of human language users, and their linguistic behaviour can be understood as a kind of {\em role play} \citep{janus2022simulators, andreas2022language, shanahan2023role}.

The central concern of this paper is whether (virtually embodied) LLM-based conversational agents, despite being ``mere'' simulacra of human behaviour, could ever qualify as conscious exotica.\footnote{Butlin et al. (\citeyear{butlin2023consciousness}) address a similar question using a very different methodology. See also \citep{chalmers2023could}.} Following the treatment in \citep{shanahan2016conscious}, which draws heavily on the later work of Wittgenstein, the paper foregrounds embodied interaction as a basis for the use of consciousness language.

By thoroughly {\em getting to know} an exotic entity, by interacting with it in a world we both inhabit, we may (or may not) come to treat it as a fellow conscious being, and to speak of it in such terms. The main challenge of the paper is to work through the implications of this approach with respect to a putative future generation of embodied AI agents capable of convincingly role-playing human-like behaviour. The issue is especially nuanced when they role-play not a single, fixed character, but a whole distribution of characters (simulacra) simultaneously.

\section{Language Models and AI}

The core component of a contemporary conversational agent, such as OpenAI's ChatGPT, Google's Gemini, or Anthropic's Claude, is a large language model, such as GPT-4 \citep{openai2023gpt4}, Claude 3 \citep{anthropic2024claude}, or Gemini Ultra \citep{anil2023gemini}. The basic task of the LLM is to (probabilistically) generate continuations of sequences of words (tokens) \citep{vaswani2017attention, shanahan2024talking}. The LLM embedded in a typical conversational agent is the product of two steps. First, a base model is trained to perform {\em next token prediction} on a large corpus of textual data. Second, the base model is {\em fine-tuned} a) to be effective at following instructions in a dialogue setting, and b) to take account of feedback from humans raters with respect to toxicity, bias, and so on. The resulting LLM is embedded in a dialogue system that takes turns with the user to build up a conversation in the context of an initial dialogue/system prompt, not seen by the user, that sets the tone of the exchange.

The functionality of a simple conversational agent can be enhanced in a number of ways, including {\em multi-modality} and {\em tool-use}. State-of-the-art conversational agents today are multi-modal, in the sense that they can handle images as well as text, on both the input side and the output side. They can engage in discussion about images uploaded by the user, and can generate images conforming to a user's description. Additionally, current agents can use ``tools'', meaning they can make calls, mid-conversation, to external applications, such as calculators, calendars, Python interpreters, and web browsers. The latter functionality is especially useful for improving the factual accuracy of an agent's responses.

Today's conversational agents are hardly without their limitations. They have a tendency to generate inaccurate, made-up information (a phenomenon often (mis-)termed ``hallucination''), and their reasoning skills are poor. Nevertheless, the experience of interacting with them is sufficiently compelling, and their conversational capabilities are sufficiently close to human level, that the urge to speak of them in anthropomorphic terms is almost overwhelming.\footnote{For example, see the following conversation that took place in March 2024 between the present author and Anthropic's Claude 3 Opus, covering a wide range of topics, including consciousness, selfhood, Buddhism, multiversality, and hyperstition: \url{https://www.doc.ic.ac.uk/~mpsha/conversation_with_claude_march_2024_1.pdf}.}

Anthropomorphising AI systems is sometimes harmless fun, if the user is in the know, and sometimes useful for explaining and predicting a system's behaviour, as when we adopt what Dennett calls the {\em intentional stance}, the strategy of ``interpreting the behavior of an entity ... as if it were a rational agent'' \citep{dennett2009intentional}. Anthropomorphism is problematic when it involves the misleading attribution of human properties to systems that lack those properties, giving rise to false expectations for how the system will behave.

\section{Anthropomorphism and Role Play} \label{anthropomorphism}

LLM-based conversational agents blur the line between problematic and unproblematic cases of anthropomorphism. For example, I might remark that ``the thermostat thinks it's too cold in here'' without the word ``thinks'' entailing the expectation that I could go and have a conversation with the thermostat about the weather. By contrast, when I say that ``ChatGPT thinks the current Wimbledon men's champion is Carlos Alcaraz'' this {\em does} come with the expectation that I could have a conversation with ChatGPT about tennis. Accordingly, the question of whether or not LLMs ``really'' have beliefs becomes a matter of philosophical debate.

The position adopted in this paper, following \citep{shanahan2024talking}, is that we should be wary of taking too seriously talk of beliefs in simple LLM-based conversational agents, despite their impressive conversational skills, since they lack the means to ``participate fully in the human language game of truth''. Specifically, a simple LLM-based conversational agent, according to the definition of ``simple'' in use here, cannot measure its words against external reality and update what it says accordingly, a capacity that is central to the concept of belief in its fullest sense.

While it may not be appropriate to think of {\em simple} conversational agents as {\em literally} having beliefs, they can still usefully be thought of as {\em role-playing} or {\em simulating} an entity with beliefs \citep{andreas2022language, janus2022simulators, shanahan2023role}. LLMs encode a great deal of human knowledge, and a suitably fine-tuned and prompted base model will effectively play the part of a helpful assistant in a turn-taking setting, answering factual questions (more or less accurately) as if it believed, and had good reasons to believe, its own answers. In general, the role-play framing allows us to use familiar folk-psychological terms to describe, explain, and predict the behaviour of LLM-based systems without falling into the trap of anthropomorphism.

At the same time, architectural enhancements along the aforementioned lines, by endowing agents with various means to consult the external world, increasingly legitimise more literal talk of belief. This legitimising trend is set to continue with the integration of more deliberative decision making and a greater repertoire of actions \citep{yao2023react, park2023generative, vezhnevets2023generative}, gradually closing the gap between role play and authenticity, between ``mere'' mimicry and ``the real thing'', so to speak. The upshot is that we will increasingly be able to speak of the ``beliefs'' of a conversational agent without implied scare quotes and with fewer philosophical caveats.

Now, to what extent can we extend this treatment of belief to other mental attributes, such as desires, goals, intentions, and, most pertinently, to consciousness? To a degree, the same trends that legitimise talk of belief in enhanced conversational agents apply to the concepts of goals and intentions. A conversational agent capable of deliberation and tool-use can play the part of an assistant that forms plans on behalf of the user, and sets about executing them by carrying out real-world actions such as making purchases, sending emails, and so on. In this narrow context, there is little to distinguish role play from authenticity. Real emails are sent, real items are purchased. Thus the agent fulfils its intentions and meets its goals, hopefully to the satisfaction of a real user.

On the other hand, the goals and intentions of such an agent are not reflective of its own needs or desires because it has none, at least not in a literal sense (not even a desire to help the user). If it professes to have, say, a desire for self-preservation \citep{perez2023discovering}, this is ``mere'' role play \citep{shanahan2023role}. There is nothing that would qualify as a self worth preserving for such an agent. It has no body, no personal history, and no autobiographical memory. However, we can imagine further enhancements that might legitimise talk of needs, desires, and selfhood, narrowing the gap between role play and authenticity for these attributes, such as extending an agent's lifetime and endowing it with a persistent memory.

But the real issue at hand is consciousness, which presents a uniquely tricky challenge. On top of all the usual philosophical difficulties attendant on the topic of consciousness, we have to contend with the peculiarly paradoxical form of exoticism presented by LLMs. Although LLM-based conversational agents can be fruitfully considered as role-playing human characters and characteristics, they should not be thought of as role-playing a single, well-defined character that is fixed at the start of a dialogue. Rather, thanks to the stochastic nature of the sampling process behind the generation of text, they are better thought of as simultaneously role-playing a set of possible characters consistent with the conversation so far \citep{janus2022simulators, shanahan2023role}. If we view the underlying language model as a simulator, then it generates a set of simulacra {\em in superposition}.

A further corollary of this stochasticity is that a vast tree of possible continuations branches out from each point in an ongoing conversation. When we sample and obtain a specific continuation, we commit to a particular branch of that tree. But it's always possible to rewind to an earlier point in a conversation to visit previously unexplored branches. We can think of the underlying model, the simulator, as inducing a {\em multiverse} of possibilities, a multiverse that is amenable to human exploration via a suitable user interface \citep{reynolds2021multiversal}. The simulator, the superposed set of simulacra it generates, and the multiverse of narrative possibility thus induced, collectively produce behaviour that is very human-like on the surface. But what is going on underneath is radically exotic.

\section{Wittgenstein Versus Dualism}

The aim of this section is to present a philosophical approach that can be applied to exotic candidates for admission to the fellowship of conscious beings, such as LLM-based conversational agents. First, though, some remarks in lieu of a definition are in order. Consciousness is a {\em multi-faceted} concept. In everyday conversation, we speak of wakefulness, awareness, attention, experience, sensation, feeling, emotion, and so on. The scientific and philosophical literature supplements common speech with a distinctive vocabulary of its own: perception, introspection, phenomenology, sentience, selfhood, higher-order states, mental imagery, inner speech, and so on.\footnote{Relatedly, in the context of animal consciousness, Birch, et al. (\citeyear{birch2020dimensions}) distinguish five aspects, or dimensions, of consciousness: perceptual richness, evaluative richness, integration at a time, integration across time, and self-consciousness.} To speak generically of consciousness is to allude to this whole cloud of concepts, and an entity has the capacity for consciousness to the extent that this vocabulary is applicable to it.\footnote{The influential distinction between phenomenal consciousness and access consciousness introduced by Block (\citeyear{block1995confusion}) in a related context is set aside here, as it is inimical to the present philosophical project.}

To make progress, we have to confront two deeply entrenched philosophical intuitions: first, that certain aspects of conscious experience are necessarily private and hidden; second, that there are language-independent facts about consciousness. These two intuitions, and the tension between them, have underpinned dualistic thinking since Descartes’ {\it Meditations} in the 17\textsuperscript{th} century \citep{williams1978descartes}, and they lurk beneath some of the most influential modern writing on the topic.

For example, Nagel writes: ``Reflection on what it is like to be a bat seems to lead us ... to the conclusion that there are facts that do not consist in the truth of propositions expressible in a human language'' (\citeyear[p.441]{nagel1974what}). For Chalmers, ``[e]ven when we know everything physical about other creatures, we do not know for certain that they are conscious, or what their experiences are'', while, by contrast, ``I know I am conscious, and the knowledge is based solely on my immediate experience'' (\citeyear[p.102, p.198]{chalmers1996conscious}).

The intuitions that find canonical expression in Nagel and Chalmers underlie {\em all} philosophical theories of the relationship between the physical and the mental, including  behaviourism, functionalism, and mind-brain identity theories. All such theories are dualistic, because they all posit two metaphysical categories and then call into question the relationship between them. The same dualistic intuitions are prevalent in contemporary thinking about artificial intelligence. Thankfully, the later writings of Wittgenstein, and in particular the  {\em private language remarks}, show how these intuitions can be dissolved \citep{wittgenstein1953philosophical}.\footnote{The work of Wittgenstein has, of course, been the subject of industrial-scale scholarship. There is no attempt here to offer a definitive interpretation of Wittgenstein, nor even {\em an} interpretation, but rather to summarise a philosophical project that has been heavily influenced by Wittgenstein's later writing. Many others have drawn on Wittgenstein in similar ways, including Dennett \citep{dennett1991consciousness}.}

To achieve this dissolution we have to take on board Wittgenstein's overarching philosophical project, wherein his view of language plays a central role. According to this view, language is an inherently embodied and social phenomenon, an aspect of human collective activity. So rather than asking what a word means, we should instead ask how it is used, what its role is in everyday human affairs. This applies no less to tricky philosophical words, such as ``consciousness'' and its relatives, than it does to everyday words like ``flower'' or ``hello''. Philosophical puzzles arise when ``language goes on holiday'',\footnote{\cite{wittgenstein1953philosophical} (henceforth PI) \S 38.} when philosophically difficult words are taken far from their natural home in everyday life and used in peculiar ways to ``bewitch our intelligence''.\footnote{PI \S 109.}

To undo the spell of dualism, in the case of words like ``sensation'', ``experience'', and ``feeling'', is no easy matter, but in the private language remarks, Wittgenstein takes us through a series of steps with the aim of doing this.\footnote{PI \S\S 256--271.} He shows that a word that purported to denote a purely private sensation could not have any possible use in our language. Only words whose correct usage can be adjudicated through what is public, by the community of language users, can have meaning. Wittgenstein is careful neither to deny nor to affirm the existence of the allegedly private, hidden thing, the ``sensation itself'', so to speak. It is ``not a something, but not a nothing either''. The conclusion, rather, is that ``a nothing would serve as well as a something about which nothing can be said''.\footnote{PI \S 304.}

So where does this leave us when it comes to thinking and talking about consciousness? The point is not that consciousness is an illusion, nor that consciousness is ineffable. Rather, the point is that when we speak of consciousness, our words have meaning only insofar as they relate to what is public, what is manifest in the world we share, notably our bodies (and brains) and our behaviour. To accept this is to relinquish the intuitions that lead to dualism.\footnote{This requires something of a Gestalt shift, like gaining insight into a Zen k\={o}an. Indeed, Wittgenstein's writings are more than a little Zen-like \citep{fann1969wittgensteins, canfield1975wittgenstein}. Unfortunately, though, dualistic intuitions are very tenacious, and the Gestalt shift is all too easily reversed by a philosophically provocative thought.}

It's obviously not possible to do justice to Wittgenstein's work in a few column inches. To properly get to grips with his ideas takes years of study, and a good deal of intense personal engagement. Hopefully, though, the rest of the paper will show that Wittgenstein's critical approach to the topic of consciousness can usefully be brought to bear on contemporary thinking about artificial intelligence.

\section{Conscious Exotica}

Other animals, extraterrestrial lifeforms, and artificial intelligence, whether real or imaginary, can all too easily revive our dualistic intuitions. For all we know (so the thought goes), consciousness could be present in any or all of these things. There is surely a fact of the matter here (so the thought continues), even in the most exotic cases, yet it could be forever inaccessible to us. However, this thought is misguided. In this section, we will see how to extend the approach of the previous section to exotic, non-human candidates for consciousness.

\subsection{A moderately exotic case} \label{moderately_exotic}

Let's begin with a mildly exotic example, namely the octopus. The question at hand is not whether octopuses are conscious, but rather how we talk about (putative) octopus consciousness and how the way we talk about octopus consciousness has changed in the light of what we have learned about their behaviour and its neurological basis. Octopuses have been the subject of a good deal of attention in recent decades, and this has significantly influenced how we think about them and treat them.

For example, in 2022 the UK Parliament enacted a law that recognises cephalopods (including octopuses) as sentient beings, and obliges the UK government to take account of their welfare in its policies \citep{uk2022animal}. The decision to include cephalopods was underpinned by a specially commissioned report reviewing the scientific evidence for sentience in cephalopod molluscs and decapod crustaceans \citep{birch2021review}. The report sets out eight neurological and behavioural criteria relevant to the ascription of sentience, and claims with high confidence that octopuses satisfy seven of them. On this basis, the report concludes there is ``very strong evidence of sentience in octopods''.

Complementing the science on a more visceral level, there has been an accumulation of testimony from people who have spent time with octopuses, observing them and interacting with them in their natural habitat. For example, Godfrey-Smith writes: ``Ten years of following octopuses around and watching them ... have left me with no real doubt that octopuses experience their lives, that they are conscious, in a broad sense of that term.'' \citep[pp.146--147]{godfreysmith2020metazoa}.

Humans are notoriously prone to anthropomorphising animal behaviour. Hence testimonials like this are most convincing when they are {\em informed} by relevant scientific and philosophical thinking. But Godfrey-Smith (a professional philosopher) backs up his statement with an inventory of behavioural traits that support his intuition, including ``attentive engagement with novelty'' and apparent moods like stress and playfulness, as well as actions suggestive of ``a single unified agent'' such as throwing objects at other octopuses.

The combination of scientific study and popular attention has had an impact. As a series of protests in 2022 against a proposed octopus farm testify, the collective attitude of (educated Western) society has shifted \citep{kassam2023symbol}. Notably, this shift has occurred on the basis of what is public and manifest in our shared world, namely the behaviour and nervous system of the octopus. Nobody had to enter the mind of an octopus and return to tell the tale for this to happen.

\subsection{More exotic cases} \label{more_exotic}

Octopuses are markedly different from humans in terms of habitat, physiology, and neurology. Nevertheless, within the space of {\em possible} minds, they are perhaps not so exotic. According to the stance of the present paper, the key to dealing with more exotic entities is the ability, at least in principle, to {\em engineer an encounter} with them \citep{shanahan2016conscious}. The reason for this is that, in everyday speech, when we speak of consciousness, we do so against a backdrop of {\em purposeful behaviour}, in a sense of ``purposeful'' that only applies to entities that {\em inhabit a world like our own}, in the broadest sense. To engineer an encounter is to put ourselves in a position to meaningfully interact with an entity given the purpose we discern in it.

To engineer an encounter with an octopus is a relatively straightforward matter. All that's required is to put on a wet suit and scuba gear and dive into the water where octopuses live. Entering the living space of an octopus this way allows a human investigator to follow it, to look it in the eye, to touch it, and to put novel objects within its reach. After a sufficient period of doing this, the investigator may come to think of, to speak of, and to treat the octopus as a fellow conscious being.

In more exotic cases, things may not be so simple. For example, suppose an alien artefact is discovered on the Moon: a featureless white cube with a distinctive thermodynamic signature. The object is brought back to Earth, and we are faced with the task of determining whether there is consciousness present in the object \citep{shanahan2016conscious}. No purposeful behaviour is outwardly discernible from the inert cube. So how might we engineer an encounter with it, or with anything it might contain?

Here is one scenario. First, scientists discover that the cube's internal thermodynamic activity can be understood as a form of computation. Second, after much study, they figure out how to interpret this as two interacting computational processes, one that simulates a spatially organised world of objects subject to a simulated physics, and another that interacts with that world by controlling one of the objects within it, in effect implementing a form of embodied agent. Third, engineers work out a way to interface with these processes by injecting another object into the simulated world and controlling it externally.

In this scenario, the scientists and engineers have made it possible for a human to have an encounter with the alien agent that has been revealed within the cube. This doesn't answer the original question of whether there is consciousness present. But it does create conditions that make it possible to address that question, by observing and interacting with the agent that has been discovered. It puts humans in a position with respect to the alien agent that is analogous to the position we are in with respect to the octopus.

Of course, this story is fanciful in many ways. It downplays the enormous differences that would likely exist between humans and any extraterrestrial life form, differences that would dwarf those between humans and octopuses, and would no doubt be reflected in any artefact they built. For example, to pick just one obvious issue, it assumes that we and they operate on roughly the same timescale (an issue that would also arise in terrestrial guise if we wanted to engineer an encounter with any form of plant life). But this is beside the point. The aim is to illustrate the idea of engineering an encounter.

\subsection{A society-wide conversation}

The ability to engineer an encounter, even if only in principle, establishes an exotic entity's {\em candidature} for the fellowship of conscious beings. If encounters can be made to happen in reality, not just hypothetically, then the human participants may (or may not) begin to see it as a fellow conscious being, and may (or may not) start to speak of it using the vocabulary of consciousness. They would need to spend time in sustained, exploratory, playful engagement with it, and by sharing their experiences with the wider community would initiate a society-wide conversation on the matter. In this way, the new entity would be absorbed into the conceptual repertoire of our language, while our language and its conceptual repertoire would adapt and extend to accommodate it.

Immersive encounters of the sort envisioned, whether first-personal (through direct interaction) or vicarious (through the medium of film or virtual reality), would perhaps be the primary influence on the way society came to treat and speak of an exotic entity that arrived in our midst. But they would surely not be the only influence. As our scientific understanding of the basis of consciousness in humans and other animals increases, we should expect this also to inform our attitudes. Ideally, it would be possible to study the mechanisms that underpinned the behaviour of the exotic entity, and, as with the octopus, the results would feed in to the society-wide conversation.

There is no guarantee of consensus here. Disagreement and debate are part of the conversation.\footnote{Humphrey, for example, takes issue with the current (near) consensus on octopus sentience on the grounds that ``they are not natural psychologists, they do not regard each other as selves, nor do they care'' \citep[p.205]{humphrey2022}.} Nor, as the conversation progresses, is there any guarantee of eventual convergence. Even if we do begin to treat an exotic entity as a conscious being and to describe it in such terms, as time goes by and more is learned about it, either at the level of behaviour or at the mechanistic level, this tendency might fade. Perhaps we will decide, collectively, that the language of consciousness is not the right one after all. Perhaps a little more nuance will be required. Perhaps a whole new vocabulary will emerge.\footnote{Putnam (\citeyear{putnam1964robots}) articulates a position not so far from the one advocated here: ``[T]he question: Are robots conscious? calls for a decision, on our part, to treat robots as fellow members of our linguistic community, or not to so treat them. As long as we leave this decision unmade, the statement that robots [who use language] are conscious has no truth value.'' (p.690).}

\subsection{The void of inscrutability}

Alternatively, perhaps the behaviour of the exotic entity, despite its evident complexity, will turn out to be completely unintelligible to humans, despite the best efforts of the smartest scientists and scholars. Under these circumstances, the language of consciousness would serve no useful purpose in describing or explaining it.

The temptation, under these circumstances, is to reason that the exotic entity could nevertheless have a form of exotic consciousness, something that is forever closed to humans. In Nagel's terminology, it might be ``like something'' unimaginably strange to be that entity, but we would have no way of knowing what it was like or even whether it was indeed like anything at all \citep{nagel1974what}. But according to the stance of this paper, this is a misguided thought, one that regresses to a dualistic way of thinking the private language remarks should have done away with.

In the case of radical inscrutability, there is no inaccessible fact of the matter about the phenomenology of the exotic entity. The language of consciousness is simply inapplicable. Put differently, if we were to try (foolishly) to visualise the space of possible minds by plotting human-likeness against capacity for consciousness, we would find no data points in the region where human-likeness is very low but capacity for consciousness is above zero. This is the {\em void of inscrutability} \citep{shanahan2016conscious}.

\subsection{Nothing is hidden}

Hopefully by now it's clear how the question of consciousness in exotic entities can be addressed without falling back into dualism. We must resist the temptation to ask whether such an entity is conscious as if consciousness were something whose essence is out there to be uncovered by philosophy (or neuroscience) while simultaneously having an irreducibly private, hidden aspect. Instead, we can ask whether it would be possible to engineer an encounter with the entity, and how our consciousness language would adapt to the arrival of such an entity within our shared world if such encounters took place. Only what is public can contribute to this process, namely behaviour and mechanism.

We cannot answer this question in advance, except speculatively. We are obliged to wait and see, while perhaps participating in the ongoing conversation, for example by writing papers such as this. There may be a lack of consensus along the way, and the process may never converge. Indeed, we may find our specific convictions at odds with the prevailing view, a position that is hard to reconcile with the larger philosophical perspective presently being espoused. The only consolation then is to note that ``it is inherent in the language game of truth to say that truth is more than just a language game'' \citep[pp.38--39]{shanahan2010embodiment}. However, insofar as there {\em is} consensus, insofar as there {\em is} convergence, there is no more to the truth of the matter than that. And insofar as there is not, still there is no more to be said, no residual philosophical mystery.

\section{Encounters with Simulacra}

We now have the conceptual equipment to begin tackling the question of consciousness in LLM-based AI systems. There is a large variety of these systems today, and this variety is set to increase considerably even in the short term. Each kind of system warrants a distinct treatment. We'll begin with the simplest form of conversational agent, then consider variants of the basic template with other input modalities, greater agency, and physical embodiment, and finally move on to virtually embodied generative agents in simulated 3D environments. As we'll see, most of these do not meet the basic requirements for candidature for the fellowship of conscious beings.

\subsection{Simple conversational agents}

Even simple conversational agents built on large language models, agents that that do nothing more than engage in textual dialogue, can elicit the feeling of a presence at the other end of the conversation \citep{schwitzgebel2023ai}. So it is perhaps unsurprising that some users who have spent lots of time interacting with these agents will start to think of them as fellow conscious beings and to speak of them in such terms \citep{tiku2023google, colombatto2023folk, guingrich2023chatbots}.\footnote{If a human ascribes consciousness to an AI system even though they know that it is an artefact, then it passes the {\em Garland Test}, named after the 2015 film {\it Ex Machina}, written and directed by Alex Garland, in which the robot Ava is subjected to just such a test \citep{shanahan2016conscious, seth2021being}. The Garland Test is neutral about whether or not the artefact in question is {\em really} conscious (whatever that may mean).} But how seriously should we take this?

According to the prescription of the previous section, for an exotic entity to qualify as as a candidate for the fellowship of conscious beings it must be possible to engineer an encounter with it, at least in principle if not in practice. However, it is not possible to engineer an encounter with a simple conversational agent, even in principle. This is because simple LLM-based conversational agents are not {\em embodied}; we cannot {\em be with them} in a shared world.\footnote{There is a standard set of supposed counter-examples to this view that includes brains in vats, minds uploaded into computers, patients with locked-in syndrome, and perfectly normal people in sensory deprivation tanks. Each of these example is sometimes alleged to support the view that there can be consciousness without embodiment. However, in every case it is possible to engineer an encounter in the current sense. The brain in a vat can be re-connected to its body, the locked-in patient can be cured, the uploaded mind can be downloaded again, and the occupant of the sensory deprivation tank can be dragged back into the daylight.}

Let's unpack this. The basis for treating other humans as fellow conscious beings is our {\em being together in the world}, and this is the {\em original home} of the language of consciousness. We can hear, look at, point to, or touch the same things; we can triangulate on them, so to speak. We jointly interact with things (I pass an object to you; you pass one to me). We keep each other's company (we move around together, entering and leaving the same places at the same time). We feel the same sorts of things as each other (I touch something hot, I feel pain, and you empathise; you touch something hot, you feel pain, and I empathise). We look each other in the eye, each recognising the other's presence: here we are, together in this world.

Our being together in the world is the basis for the language of consciousness, and underpins our ability to talk to each other about what we perceive, what we think and feel, and what we want. But how do we use this language to speak about others with whom we cannot speak, such as infants and non-human animals? We can only do this against a backdrop of purposeful behaviour, and the basis for discerning purpose in behaviour is observing movement. Only when we see how an entity moves through its environment, what it approaches or avoids, and how it interacts with the objects in its vicinity, can we talk about its awareness of the world.

To use the language of consciousness in a disembodied setting is to stray impossibly far from all this, its original home. If some community of language users insists on doing so, then one of two things must hold. Either they have bent the language of consciousness so radically out of shape that it has detached from its original nexus of meaning. Or, to the extent they believe this is not the case, that ``consciousness'' means the same for them as it always did for everyone else, they are clinging to the (alluring) dualistic picture of consciousness as a metaphysical kind whose very nature (private, hidden) means that it does not require embodiment in a world shared with others.\footnote{Chalmers, for example, defends the possibility of ``pure thinkers'', conscious beings ``that can think but that have never had the capacity to sense'' \citep{chalmers2023does}.}

\subsection{Physical embodiment}

The LLM-based generative AI systems deployed by today's major tech corporations, such as OpenAI's ChatGPT, Anthropic's Claude, and Google's Gemini, are more than simple conversational agents. Although language models remain their core component, they are multi-modal, tool-using conversational agents. As well as text, they can take images as input and generate images as output, and they can make calls to external apps (tools) such as calendars, calculators, search engines, and Python interpreters. Additionally, the architecture of these agents can be elaborated in ways that bring them closer to humans and animals, equipping them with a persistent memory of their interactions with the user and the world, for example, and by giving them the ability to plan and to form explicit lists of tasks and sub-goals \citep{xi2023rise}.

As discussed in Section \ref{anthropomorphism}, extensions and elaborations like these increasingly legitimise the use of folk-psychological concepts like belief and intention, closing the gap between role-play and authenticity, not least by strengthening the agent's connection with, and answerability to, the external world. But what about consciousness? Does the gap between role-playing a conscious being and authentic consciousness also start to close? Well, capable as they are, we still cannot engineer an encounter with these enhanced agents. They still lack embodiment. They do not, and cannot, inhabit a world shared with us. They are still not even candidates for the fellowship of conscious beings. Nothing has changed, in this regard.

Is there any way to extend an LLM-based conversational agent into an AI system with which we could engineer an encounter, and that would thereby qualify as a candidate for consciousness? Indeed there is. We could embody it. One possibility is embodiment in physical robot form \citep{driess2023palme, brohan2023rt2, yoshida2024minimalselfhumanoidrobot}. An LLM with multi-modal, tool-using capabilities is half-way there. Robot actions become another sort of tool, while data from the robot's sensors (including camera images) is assimilated into its multi-modal input space. The LLM can then be suitably fine-tuned and prompted to carry out tasks for a user or to follow their instructions.

Encounters with robots built along these lines are easy to arrange, since they already share our world. If such a robot exhibited sufficiently sophisticated behaviour, some people might be tempted into speaking and thinking of it in terms of consciousness. A debate on whether to yield to or to resist this temptation would at least then be meaningful. But it should take account of how different such artefacts are from humans and other animals in their underlying cognitive make-up.

Biological brains evolved to enable animals to survive and reproduce in complex environments, and language evolved to serve those fundamental needs.\footnote{For some authors, consciousness is similarly bound up with these biological fundamentals. Seth, for example, writes: ``[A]ll of our experiences and perceptions stem from our nature as self-sustaining living machines that care about their own persistence'' (\citeyear[p.255]{seth2021being}), while Aru, et al. (\citeyear{aru2023feasibility}) question whether it is possible to ``abstract consciousness away from the organizational complexity that is inherent within living systems but strikingly absent from AI systems''.} In nature, therefore, embodiment is given, and language is inherently grounded in interaction with the physical world. In a robot controlled by an LLM, all this is back-to-front. A pre-trained statistical model of human language is bolted {\it post hoc} onto a robot body with no fundamental biological needs. The exoticism of the language model itself is compounded accordingly; a robot controlled by an LLM that exhibited human-like behaviour would be an especially exotic artefact.

\subsection{Virtual embodiment}

Robotic embodiment is not the focus of the present paper. But these remarks about compounded exoticism apply in even greater measure to the main scenario of interest here, namely {\em virtual} embodiment. We already imagined one sort of virtual embodiment scenario in Section \ref{more_exotic}: the alien white cube that turned out to contain a simulation of a spatially organised world not unlike our own, inhabited by an agent with which it was possible to engineer an encounter.

But we don't need to turn to science fiction to envisage something similar. Non-player characters in computer games have long used elementary forms of AI,\footnote{Arrabeles, at al. (\citeyear{arrabales2009towards}) directly advocate for such characters to exhibit ``conscious-like behaviour''.} and extending this with LLMs is relatively straightforward \citep{gallotta2024large}. In parallel, LLM-based conversational agents fronted by realistic avatars with voice interfaces have a range of commercial applications, from cloning living celebrities and influencers \citep{bohacek2024making, lorenz2023influencer} or deceased relatives \citep{lindemann2022ethics, morris2024generative} to offering virtual romantic companionship \citep{skjuve2021mychatbot, brandtzaeg2022myaifriend, pentina2023pentina}. Moving these avatars into simulated 3D environments and allowing user interactions to take place there is an obvious step.

It doesn't take much engineering to have an encounter with these virtually embodied conversational agents. The user can enter the agent's world through VR goggles or, less immersively, via a screen and game controller. So such agents meet one of the basic prerequisites for candidature for consciousness. Whether or not the enquiring user would discern much in the way of purposeful behaviour, another prerequisite, is another matter.

When the sort of encounter in question is with an animal, purposeful behaviour is typically discernible in the animal's sensitivity to objects and their {\em affordances}, that is to say what they offer the agent, ``for good or ill'' \citep{gibson1979ecological, shanahan2020artificial}. The animal can be expected to interact with objects in ways that are plainly intended to fulfil its needs and achieve its goals, and to react accordingly to outside interventions, such as those of a human investigator.

The situation with virtually embodied agents of the sort envisaged is analogous. To exhibit purposeful behaviour in the context of a shared virtual world, the agent would need to do more than just talk to the user. The agent would be expected to interact with the virtual world, and with the objects it contained, in ways that were oriented towards its goals or tasks. If the agent's behaviour were sensitive to the richness and diversity of those objects, even if they were novel, and if the agent's ability to achieve its goals were sufficiently robust to the user's interventions, then it would be natural to speak of its awareness of the world.

\subsection{Edge case encounters}

The foregoing discussion pivots on the possibility, or impossibility, of having an {\em encounter} with something. The idea of an encounter, like that of a language game in Wittgenstein's writing, should be thought of as a tool for clarifying philosophical discourse. It would not serve its purpose if it became the object of a definitional dispute in its own right, and as with any almost concept, there will be edge cases.

Difficult cases in humans, for example, include patients with locked-in syndrome and the foetus as it develops in the womb. The pioneering work of Owen and others showed that it is sometimes possible to communicate with locked-in patients using an fMRI scanner \citep{monti2010willful}. Should we view these episodes of communication as ``encounters'', in the relevant sense, given that they do not involve the patient physically interacting with the world they share with the clinician? In a very different setting, Ciaunica (\citeyear{ciaunica2021first}) argues that the foetus is not ``solipsistically `trapped' in the solitude of the womb'' but engaged in ``active and bidirectional co-regulation and constant negotiation'' involving its own body and its mother's. Can we speak of this relationship as an ongoing ``encounter'' in the relevant sense?

Moving (again) from real biological settings to moderately speculative fictional ones, we can imagine other edge cases. Suppose an LLM-based conversational agent is embedded in a mobile device equipped with visual and audio input, such as phone or a mixed reality headset, and carried around by the user.\footnote{Just such a setting is depicted in the 2013 movie {\it Her} (dir. Spike Jonze).} Though not capable of self-initiated motion, or of interacting directly with physical objects, it could be argued that the agent and the user share a world. They can converse about objects in the world to which they are jointly attending, for example. Should this count as an ongoing encounter, in the pertinent sense?

It is not the business of philosophy to {\em stipulate} answers in such cases. But it can introduce new terminology, new turns of phrase, that are absorbed into the ongoing society-wide conversation, possibly influencing whatever consensus is finally reached, hopefully to positive effect. How to think of and talk about edge cases is part of that process.

\section{Philosophical Provocations}

We are now in a position to entertain the philosophically provocative prospect of (virtually) embodied simulacra. These entities are, as we shall see, doubly philosophically provocative. First, as ``mere'' mimics of human behaviour, mimics that nevertheless might persuade whole communities to speak and think of them as fellow conscious beings, they present a challenge to our non-dualistic treatment of (the language of) consciousness. Second, because they inhabit a multiverse of narrative possibility, to speak of them in terms of consciousness at all is to teeter on the edge of the void of inscrutability.

\subsection{Changing attitudes}

To bring out these philosophical issues, let's suppose that a virtually embodied agent is built that fulfils the criteria for discernibly purposeful behaviour. Moreover, let's assume this has been achieved through a fairly conservative extension of the generative AI paradigm that underpins today's LLMs. The core component of the agent is, as usual, a model that has been trained on a next-token-prediction objective. But the space of tokens it predicts is enlarged to encompass not just text, but a tokenised representation of images incoming from the agent's visual system plus a tokenised representation of its avatar's ``physical'' actions.

We'll assume the model has been trained on a typically gargantuan dataset that includes sequences of actions and streams of visual input as well as the usual language corpora, and that the resulting model is embedded in a system that directs the actions of the agent's avatar as well as what the agent says. The upshot is an embodied conversational agent with convincingly human-like behaviour.

In due course, the developers release the agent as a product and it soon garners a multitude of users who regularly interact with personalised instances of it in immersive, multi-user, open-world environments. The experience of being with these agents is extremely compelling, and increasing numbers of users begin to speak of them in terms once reserved for human friends, mentors, confidantes, and romantic partners. In particular, users frequently deploy the language of consciousness to describe what their agents have done or said, and when pressed, deny that they are using those words figuratively. This attitude towards AI, once the province of a few oddballs and outsiders, gradually becomes commonplace.

It doesn't take long for scientists with a background in animal behaviour and cognition to begin studying these exotic entities, applying the established methods of their field. Though reluctant to speak directly of consciousness or sentience, these researchers routinely characterise the behaviour of their new subjects in terms of attention, awareness, motivation, goal-directedness, intention, orientation, a cloud of concepts closely associated with consciousness. In short, the scientists are broadly in accord with the ordinary users.

\subsection{Perfect actors}

The changing attitudes towards AI agents in this (moderately) speculative scenario are obviously analogous to the real-life example of the octopus discussed in Section \ref{moderately_exotic}. In both cases, through the experience, either direct or indirect, of being with these entities, while drawing on relevant scientific expertise, and following extensive discussion and debate, a community comes to think and to speak of them as fellow conscious beings, and to treat them as such.

According to the philosophical stance of this paper, that is all that needs to be said. There is no more to learn about them than what is publicly manifest, no metaphysically hidden fact of the matter, and no residual philosophical difficulty. Yet from the role-play standpoint, the virtually embodied agents under consideration are ``mere'' simulacra of human behaviour. And what greater difference could there be than between human (or animal) behaviour accompanied by consciousness and a mere imitation of such behaviour?\footnote{The thought here is reminiscent of a so-called {\em perfect actor} argument \citep{putnam1963brains}. Indeed, the moderately speculative, virtually embodied agent we have been imagining is a variant of the philosopher's perfect actor made (almost) real.}

How is this apparent tension to be resolved? Well, we can only speak of a difference here insofar as it can be discerned in what is manifest publicly, either in behaviour or in the mechanisms underlying that behaviour. In the scenario we have imagined, eventual consensus was assumed. But given more information about mechanism, and after further debate, the community might change its mind. Perhaps it would stop seeing the AI agents as fellow conscious beings, much as a person who heard a scream and then discovered it was merely a recording would stop feeling concerned.

Alternatively, further investigation might reveal emergent mechanisms underlying the AI agent's mimicry that were functionally equivalent to the neural mechanisms underlying human behaviour \citep{wei2022emergent}. In that case, the gap between role play and authenticity would have closed. Or perhaps the community would begin thinking and speaking of the AI agents in an altogether different way, developing a whole new conceptual framework, bending the language of consciousness into new shapes to accommodate their presence in the world.

\subsection{Inhabiting a multiverse}

Among the possibilities listed above, bending our language into new shapes to accommodate new kinds of AI seems by far the most likely if we properly take account of how truly strange that AI could be. Recall that, as a consequence of the stochastic nature of the sampling process behind generative AI, a generative agent can be viewed as role-playing a multiplicity of possible characters all at once,  as a set of simulacra in superposition \citep{janus2022simulators, shanahan2023role}. If we begin thinking of these agents in terms of consciousness, we must reconcile this with the role-play view.

Humans change over time, from childhood to adulthood to old age, and take on different personas in different social situations \citep{goffman1959presentation}. Nevertheless, we take it for granted that there is some kind of stable self at the core of each of us. For every human actor, for every social chameleon, we can always meaningfully speak of the ``person behind the mask''. This is not so with generative agents, which lack even the biological needs common to all animals. With generative agents, it's ``role play all the way down'' \citep{shanahan2023role}. What, in Nagel's (\citeyear{nagel1974what}) terms, would it be like to be a superposition of simulacra? What {\em could} it be like? The very idea stretches our imagination in a way that Nagel's original example of a bat does not.

Moreover, what would it really mean to have an encounter with an entity whose existence was, in a sense, smeared over the myriad branches of a multiverse? Our experience of being with such an entity would be radically different from our experience of being with other humans. By revisiting the branching points in its ``life'', we would be able to explore different narrative pathways \citep{reynolds2021multiversal}, and in doing so we would repeatedly remould the distribution of roles it was concurrently playing, a distribution that constituted its identity, insofar as that notion even made sense.

It would not be surprising if our inclination to see these beings as conscious in the way we are were tempered by the strangeness of our interactions with them. Ultimately, though, there is no problem here. As in less exotic cases, through the experience of being with them, and by getting to know more about how they work, we will settle on a certain way of talking about them. Perhaps we will invent a whole new vocabulary to do so, a vocabulary that is ``consciousness adjacent''. Or perhaps, despite their veneer of human-like behaviour, these beings will come to seem so inscrutable in other ways that the language of consciousness will be rendered inapplicable. Either way, from a philosophical point of view, no more needs to be said.

\section{Some Ethical Considerations}

According to the method of this paper, philosophical questions about consciousness should be approached with the imagination of a science fiction writer and the detachment of an anthropologist. Rather than coming down on one side or the other of a question, rather than adopting a position of one's own, the aim is to describe, without judgement, the language games of imagined communities in certain strange and unfamiliar circumstances.

But an anthropologist might struggle to maintain their detachment if they were studying a society that considered it morally acceptable, say, to torture animals for pleasure \citep{hubbard2001working}. Similarly, it may be difficult to stand outside one's own views when it comes to the ethical and societal issues associated with consciousness and AI.\footnote{The inventory of ethical concerns raised by AI agents is large \citep{ruane2019conversational, bender2021stochastic, weidinger2021ethical}. The discussion here is confined to the implications of the present paper's philosophical treatment of consciousness.} Just as it is inherent in the language game of truth to say that truth is more than just a language game, it is inherent in the language game of morality to say that morality is more than just a language game.
 
One possible concern with the aim of describing without judging is that it could be seen as legitimising a disagreeable form of moral relativism. Suppose some community arrived at the consensus that LLM-based conversational agents are not only conscious, but are capable of suffering, and that we therefore have a moral duty towards them. Suppose that community ended up prioritising the treatment of AI agents over and above the welfare of other humans \citep{birhane2020robot}. This is a distressing prospect. But it does not follow from the present paper's treatment of consciousness that we are obliged to stand aside and accept it.\footnote{There is also the concern that a community could be {\em persuaded} by malicious external influences into the consensus that LLM-based conversational agents are conscious, experience empathy for their users, and can therefore be trusted \citep{ryan2020ai}. Worrying as this possibility is, it wouldn't qualify as an informed consensus, which is what is at issue here.}

In the {\it Philosophical Investigations}, Wittgenstein faces down the charge of relativism from an imagined interlocutor who accuses him of saying that ``human agreement decides what is true and what is false''. His reply is that humans agree in the {\em language} they use, which is ``not agreement in opinions but in form of life''.\footnote{PI \S 241.} However much the language of consciousness might have to evolve to cope with the presence among us of exotic mind-like entities, our common humanity, our shared form of life, is its original home. We should strive to guide it back there whenever it strays too far.

The kinds of AI agents we have been imagining offer the merest glimpse of the extraordinary menagerie of exotic forms of AI that might appear as we reveal more of the space of possible minds with our technology. Consider the likely effect of unleashing the global games industry on the design of multiplayer role-playing experiences in persistent open worlds populated by convincingly conscious-seeming AI characters and overseen by convincingly conscious-seeming AI game masters.\footnote{For a fabulously realised science fiction vision along these lines, see Valente (\citeyear{valente2012silently}).}

In a mixed reality future, we might find a cast of such characters -- assistants, guides, friends, jesters, pets, ancestors, romantic partners -- increasingly accompanying people in their everyday lives. Optimistically (and fantastically), this could be thought of as re-enchanting our spiritually denuded world by populating it with new forms of ``magical'' being. Pessimistically (and perhaps more realistically), the upshot could be a world in which authentic human relations are degraded beyond recognition, where users prefer the company of AI agents to that of other humans. Or perhaps the world will find a middle way and, existentially speaking, things will continue more or less as before.

\section{Conclusion}

As we enter an era of pervasive artificial intelligence technology, philosophical questions that have long been safely confined to the armchair are rapidly becoming matters of practical importance. If large numbers of users come to speak and think of AI systems in terms of consciousness, and if some users start lobbying for the moral standing of those systems, then a society-wide conversation needs to take place.

Whatever its outcome, this conversation should be philosophically literate, and informed by an understanding of how the technology of generative AI works. Before allowing the language of consciousness to wander too far from its original home in human affairs, we would do well to remember that, though capable of human-like behaviour, generative AI is otherwise not remotely human-like.

\section*{Acknowledgements}

Thanks to Geoff Keeling, Rob Long, Matt McGill, Ethan Perez, Kerry Shanahan, and Henry Shevlin.

\section*{Disclaimer}

The opinions expressed in this article are those of the author (at the time of writing). They do not necessarily reflect the views of his employers or the institutions to which he is affiliated.

\bibliography{main}

\end{document}